\pdfoutput=1
\documentclass[11pt]{article}

\usepackage[]{EMNLP2022}

\usepackage{times}
\usepackage{latexsym}
\usepackage{diagbox}
\usepackage{dashrule}
\usepackage{arydshln}
\usepackage[T1]{fontenc}
\usepackage{pifont}
\newcommand{\cmark}{\ding{51}}%
\newcommand{\xmark}{\ding{55}}%

\usepackage[utf8]{inputenc}

\usepackage{microtype}

\usepackage{inconsolata}

\usepackage{graphicx}
\usepackage{multirow}
\usepackage{array,booktabs}
\usepackage{url}

\newcommand*\samethanks[1][\value{footnote}]{\footnotemark[#1]}

\newcommand{\eg}{\textit{e}.\textit{g}. }

\title{Rethinking the Authorship Verification Experimental Setups}

\author{Florin Brad\thanks{\hspace{0.5em}Equal contribution.} \\
  Bitdefender \\
  \\\And
  Andrei Manolache\samethanks \\
  Bitdefender,\\ University of Stuttgart \\ 
  \texttt{\{fbrad,amanolache,eburceanu\}@bitdefender.com} \\\And
  Elena Burceanu \\
  Bitdefender \\
  \\\AND
  Antonio Barbalau \\
  University of Bucharest \\
  \texttt{abarbalau@fmi.unibuc.ro} \\\And
  Radu Tudor Ionescu \\
  University of Bucharest \\
  \texttt{raducu.ionescu@gmail.com} \\\And
  Marius Popescu \\
  University of Bucharest \\
  \texttt{popescunmarius@gmail.com} \\
  }
  
\begin{document}
\maketitle
\begin{abstract}
One of the main drivers of the recent advances in authorship verification is the PAN large-scale authorship dataset. Despite generating significant progress in the field, inconsistent performance differences between the closed and open test sets have been reported. To this end, we improve the experimental setup by proposing five new public splits over the PAN dataset, specifically designed to isolate and identify biases related to the text topic and to the author's writing style. We evaluate several BERT-like baselines on these splits, showing that such models are competitive with authorship verification state-of-the-art methods. Furthermore, using explainable AI, we find that these baselines are biased towards named entities. We show that models trained without the named entities obtain better results and generalize better when tested on DarkReddit, our new dataset for authorship verification.
\end{abstract}

\section{Introduction}

Identifying the author of a text is one of the most versatile NLP tasks, with applications ranging from plagiarism detection to forensics and monitoring the activity of cyber-criminals. The task spans several decades and was tackled using statistical linguistics \cite{mendenhall1887,zipf1932selected,MostellerWallace64}, and, more recently, machine learning \cite{svmmail,dt,Koppel-JMLR-2007,Stamatatos2009}. Due to the typically small data setup of authorship analysis tasks, deep learning methods had a slow start in this domain. Nevertheless, inspired by the impressive performance of pre-trained language models, such as BERT \cite{Devlin2019}, these methods gained traction in authorship analysis as well. \newcite{SAEDI2021101241} showed that Convolutional Siamese Networks are more robust than a BERT-based method over large-scale authorship attribution tasks.

\newcite{Barlas2020CrossDomainAA} investigated pre-trained language models for cross-topic and cross-domain authorship attribution and showed that BERT and ELMo \cite{Peters:2018} achieve the best results while being the most stable approaches. \newcite{fabien-etal-2020-bertaa} introduced BERT for Authorship Attribution (BertAA) in which they combine BERT with stylometric features for authorship attribution. The authors remarked that their model is unable to perform text similarity evaluation in the context of the more difficult authorship verification problem, which we tackle.

One of the main contributors to the active developments in authorship analysis is the PAN organizing team, who proposed annual shared tasks since 2009. While the recent PAN 2020 and 2021 contests increased the difficulty of the authorship verification task and enabled large-scale model training \cite{Kestemont2020OverviewOT,kestemont2021}, there are still possible generalization issues due to the dataset splits. For instance, models from 2020 trained on the \emph{closed-set} data surprisingly performed better on the \emph{open-set} test data (which is arguably more difficult) than on the \emph{closed-set} test data \cite{kestemont2021}. We therefore argue that in order to better assess the generalization capabilities of authorship verification systems, a more fine-grained approach to dataset splitting may be needed. 
To address these issues, we introduce a set of five carefully designed splits of the publicly available PAN dataset, ranging from the easiest setup (\emph{closed-set}) to the most difficult (\emph{open-set}). Our splits progressively alleviate information leaks in the test data, enabling a more confident evaluation.

Furthermore, we release our splits publicly\footnote{\url{https://github.com/bit-ml/Dupin/tree/main}} to allow other members of the community to evaluate models on any computing infrastructure, enabling the evaluation of large-scale models. Along with the new splits, we introduce a set of BERT-based models \cite{Devlin2019} to serve as baselines for future research. We show that these language models are competitive with the top scoring O2D2 (\textit{out-of-distribution detector}) system at PAN 2021 \citep{boenninghoff:2021}.

We also qualitatively inspect the models' predictions and find that they often rely on named entities to verify authorship. We show that by replacing the named entities in the dataset with placeholders, we are able to obtain significant performance gains and better generalization capabilities.

\begingroup
\begin{table}[t]
    \setlength{\tabcolsep}{1.4pt} % Default value: 6pt
    \begin{center}
        \begin{tabular}{ l  ccccc }
            \toprule
            \shortstack{Test split} &
            \shortstack{\textit{O2D2\textsuperscript{$\ast$}}} &
            \shortstack{\textit{O2D2}} &
            \shortstack{\textit{BERT}} &
            \shortstack{\textit{Naive\textsuperscript{$ \dagger $}}} &
            \shortstack{\textit{Comp.\textsuperscript{$ \dagger $}}} \\
            
            \midrule
            Closed   & 93.5 & \textbf{96.4} & 95.6 & 75.6 & 72.2 \\
            Clopen  & 94.0 & 96.0 & \textbf{97.4} & 74.1 & 71.1 \\
            Open UA & 92.6 & \textbf{92.6} & 90.2 & 78.6 & 68.5 \\
            Open UF & 91.4 & \textbf{95.1} & 91.6 & 79.9 & 79.0 \\
            Open All & 80.6  & 67.5 &\textbf{ 88.7} & 75.6 & 76.9 \\ 
            \hdashline\noalign{\vskip 0.5ex}
            PAN Closed & 93.3 & \textbf{93.5} & - & 74.7 & 74.2 \\
            PAN Open & 93.3 & \textbf{94.4} & - & 75.3 & 74.5  \\

            \bottomrule
        \end{tabular}
    \end{center}
    \caption{Overall scores of several models evaluated on our public test splits. We also list the reported results of the models on the private PAN splits. BERT is competitive with the top-scoring O2D2 model of the PAN 2021 competition and both methods greatly outperform the PAN baselines (Naive and Compression). O2D2 performs poorly on our most difficult split Open All. However, performance on the development set is much closer to the BERT results on the test set. The neural models were trained on the large training splits. $^\dagger$Models trained on the small datasets. $^\ast$Models evaluated on the validation set.  }
    \label{tab: baselines}
\end{table}
\endgroup

In summary, \textbf{our contributions} are threefold:

\textbf{1.} We \textbf{introduce five splits}, based on the PAN dataset, with a decreasing degree of shared information between train and test sets. These configurations enable benchmarking large models, providing a robust evaluation environment, on which we run several BERT-based baselines.

\textbf{2.} Using explainable AI (XAI) methods, we find that \textbf{BERT-like models focus on named entities}  to determine authorship. We replace them with placeholders and retrain our models, which brings a significant performance boost.

\textbf{3.} We introduce the DarkReddit dataset for authorship verification, which is significantly different in style to the fanfictions in PAN. We test the \textbf{generalization capabilities} of the models trained on PAN, by evaluating them on DarkReddit. % in a zero-shot scenario. 
Our previous finding is further confirmed by our model trained without named entities, which generalizes better and improves the overall metric by 5.6\%.

\begingroup
\begin{table}[t]

    \setlength{\tabcolsep}{0.3pt} % Default value: 6pt
    \begin{center}
        \begin{tabular}{ l  c c c c}
            \toprule
            Split & \rotatebox[origin=c]{70}{authors in val} & \rotatebox[origin=c]{70}{fandoms in val} & \rotatebox[origin=c]{70}{authors in test} & \rotatebox[origin=c]{70}{fandoms in test} \\
            \midrule
            Closed & \cmark & \cmark & \cmark & \cmark\\
            Clopen$^{\diamond}$ & \cmark & \cmark & \cmark & \cmark\\
            Open Unseen Authors & \xmark & \cmark & \xmark & \cmark\\
            Open Unseen Fandoms & \cmark & \xmark & \cmark & \xmark\\
            Open All & \xmark & \cmark & \xmark & \xmark\\
            \bottomrule
        \end{tabular}
    \end{center}
    \caption{Dataset splits sorted from the easiest (train authors and fandoms are seen in the validation and test sets) to the most difficult (train authors and fandoms are not found in the test set). $^{\diamond}$Some of the authors of the Different Authors train pairs in Clopen may be unknown at test time, making it a mix between Closed and Open.}
    \label{tab: splits}
\end{table}
\endgroup

\begin{table*}[t!]
    \setlength{\tabcolsep}{2.8pt} % Default value: 6pt
    \begin{center}
        \begin{tabular}{l  ccc | ccc | ccc | ccc | ccc}
            \toprule &
            \multicolumn{3}{c}{Closed$_{XL}$} & \multicolumn{3}{c}{Clopen$_{XL}$} & \multicolumn{3}{c}{Open UA$_{XL}$} & \multicolumn{3}{c}{Open UF$_{XL}$} &
            \multicolumn{3}{c}{Open All$_{XL}$}\\
            \cmidrule(lr){2-4}
            \cmidrule(lr){5-7}
            \cmidrule(lr){8-10}
            \cmidrule(lr){11-13}
            \cmidrule(lr){14-16}
            \textbf{Metric} & 
            \textbf{O2D2} & \textbf{cB} & \textbf{B} &
            \textbf{O2D2} & \textbf{cB} & \textbf{B} &
            \textbf{O2D2} & \textbf{cB} & \textbf{B} &
            \textbf{O2D2} & \textbf{cB} & \textbf{B} &
            \textbf{O2D2} & \textbf{cB} & \textbf{B}\\ 
            \midrule
            \textit{F1} & \textbf{96.6} & 93.8 & 95.0 & 96.1 & 95.6 & \textbf{96.8} & \textbf{93.6} & 85.4 & 89.2 & \textbf{95.2} & 88.6 & 90.8 & 45.0 & 74.8 & \textbf{86.9} \\
            \textit{F0.5}& 94.3 & 93.4 & \textbf{94.5} & 94.0 & 96.5 & \textbf{97.0} & \textbf{89.6} & 87.0 & 88.1 & \textbf{94.5} & 92.3 & 88.8 & 70.1 & 84.6 & \textbf{87.2} \\
            \textit{c@1} & \textbf{95.9} & 93.3 & 94.5 & 95.4 & 95.4 & \textbf{96.6} & \textbf{91.5} & 84.8 & 88.2 & \textbf{93.1} & 88.8 & 90.0 & 65.2 & 78.9 & \textbf{87.0} \\
            \textit{AUC}  & \textbf{98.7} & 98.0 & 98.6 & 98.4 & 99.1 & \textbf{99.4} & \textbf{95.8} & 92.4 & 95.3 & \textbf{97.6} & 96.5 & 96.9 & 89.5 & 91.1 & \textbf{94.0} \\
            \textit{overall}  & \textbf{96.4} & 94.7 & 95.6 & 96.0 & 96.7 & \textbf{97.4} & \textbf{92.6} & 87.4 & 90.2 & \textbf{95.1} & 91.5 & 91.6 & 67.5 & 82.3 & \textbf{88.7}\\
            \bottomrule
        \end{tabular}
    \end{center}
 \caption{Comparison of neural models on the PAN 2020 XL splits. O2D2 outperforms BERT (\textbf{B}) on three out of five splits, while BERT outperforms charBERT (\textbf{cB}) on all the splits. Note how the \emph{closed-set} results (left side) are considerably better than the \emph{open-set} ones (right side), indicating that models overfit the styles of the known authors from the closed splits. We report all the PAN 2020 metrics for the test split. The best result per split is in bold.}
    \label{tab: pan2020_pretraining}
\end{table*}

\section{Datasets}
\label{sec: dataset}

We use the PAN 2020 authorship verification dataset\footnote{\url{https://pan.webis.de/clef21/pan21-web/author-identification.html}}. A document $d_i$ belongs to a fandom (topic) $f_i$ and is written by an author $a_i$. Author verification is a classification task which asks whether documents $d_i$ and $d_j$ are written by the same author (SA) or by different authors (DA). The dataset comes in two sizes: small (52k examples) and large (275k examples). The latter one is better suited for deep learning models.

\subsection{New PAN 2020 splits}
The PAN 2020 competition is a \emph{closed-set} verification setup, meaning that the unseen test set contains documents whose authors and fandoms were seen at training time. The PAN 2021 competition has a more difficult \emph{open-set} setup, in which the training data is the same as in 2020, but the submitted solutions are privately tested against document pairs from previously unseen authors and fandoms. 
The PAN testing infrastructure makes it difficult to evaluate large models quickly. To this end, we release several dataset splits, ranging from the easier \emph{closed-set} setup to the more difficult \emph{open-set} variants. We summarize the splits in Tab.~\ref{tab: splits} and provide a more detailed description in the Supplementary Material~\ref{sec: apx_pan_splits}. For each split, we propose a small (XS) and a large (XL) version.

\subsection{DarkReddit}
\label{sec: darkreddit_dataset}
\begin{figure*}[t!]
    \begin{center}
        \includegraphics[width=0.99\textwidth]{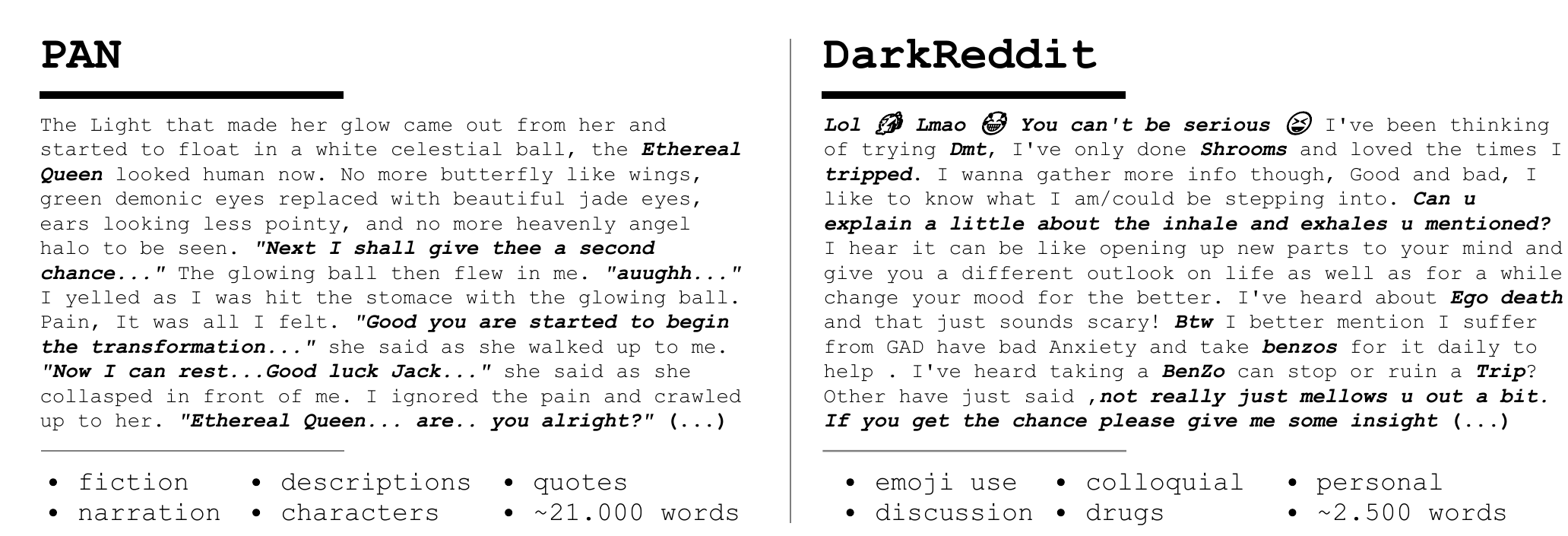}
    \end{center}
    \caption{A PAN-2020 sample compared to a DarkReddit one. Note the contrasting style, topics, vocabulary and size between the two samples.}
    \label{fig:pan_reddit_samples}
\end{figure*}

To test an even more difficult scenario than our \emph{open-set} splits, we created a small authorship verification dataset. This dataset could be used to benchmark the generalization capabilities of AV models, while also being useful for cybersecurity applications. The dataset was constructed by crawling $1026$ samples from $\texttt{/r/darknet}$\footnote{\url{https://www.reddit.com/r/darknet/}}, a subreddit dedicated to discussions about the Darknet. There is an equal number of same author and different author pairs, resulting in a balanced dataset. A document has 2,500 words on average, $9$ times less than the PAN 2020 splits. The two datasets also differ in other aspects (\eg$\!$topics, authors, text purpose, self-contained message). We illustrate the differences between PAN and DarkReddit examples in Figure~\ref{fig:pan_reddit_samples}.

\section{Experiments}
\noindent\paragraph{Training.} We fine-tune \textbf{BERT} (\textbf{B})~\cite{Devlin2019} and \textbf{Character BERT} (\textbf{cB, charBERT})~\cite{Boukkouri2020} as binary classifiers for authorship verification. Given two documents $d_i$ and $d_j$, we concatenate and feed them to the Transformer encoder. When a document is longer than 256 tokens, we sample a random chunk of length 256. The chunks are resampled at every epoch, hence increasing the variety of the training set. To make predictions, we add a linear layer on top of the $h_{[CLS]}$ vector and optimize the entire model via the binary cross entropy loss. We use the same set of hyperparameters across all of the experiments. For the other models (\textbf{O2D2}, \textbf{Naive} and \textbf{Compression}) we used the provided code and default hyperparameters.

\noindent\paragraph{Evaluation.} We report the $overall$ metric from PAN 2020 (the mean over $F1$, $F0.5$, $c@1$ and $AUC$). To use information from all document pairs $(d_i, d_j)$, we split each of them into 256-length non-overlapping chunks. We then feed each chunk pair to the model, obtaining the class probabilities. Finally, we average the probabilities of all the chunk pairs to obtain the prediction for the document pair.
Unsurprisingly, using multiple chunks outperforms randomly picking only one chunk from each document, leading to up to 10\% improvements in the overall score.

% \section{Results}

\begin{figure}[t!]
    \begin{center}
    \includegraphics[width=0.49\textwidth]{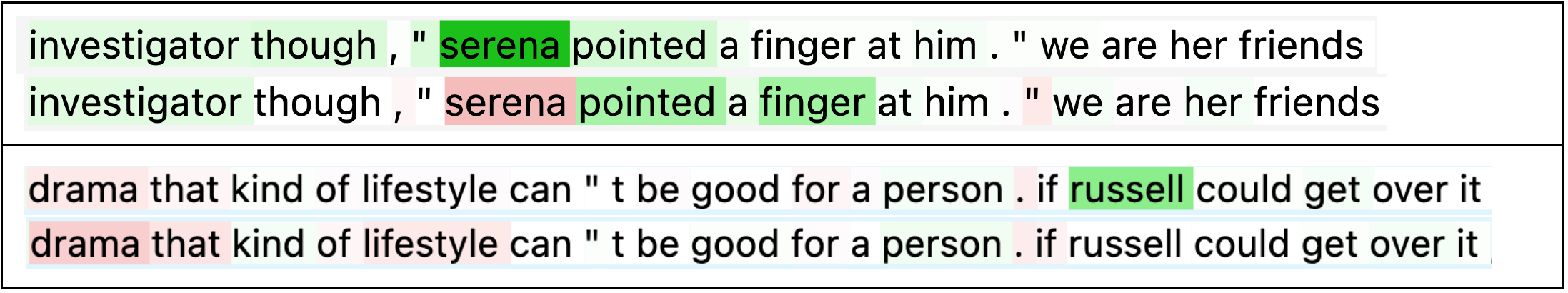}
    \end{center}
    \caption{
    Explainability analysis using Integrated Gradients. Words highlighted in green help the correct prediction, while those in red distract from it. In each pair, the rows' attributions are from BERT fine-tuned on Closed$_{XL}$ and Open UF$_{XL}$ respectively. Fine-tuning on the latter split changes the words' attribution scores. Specifically, the focus on named entities (\eg serena, russell) in the 1st row of each pair, which should not be relevant in author detection, diminishes in the 2nd row.
    }
%for each pair, the first row represent the model fine-tuned on closed XL
\label{fig: xai_on_splits}
\end{figure}

\subsection{Model comparison}
\noindent\paragraph{Comparison to PAN models.}
As can be seen in Tables~\ref{tab: baselines} and \ref{tab: pan2020_pretraining}, BERT is competitive with the PAN 2021 winner on our public test splits. Both models greatly outperform the PAN baselines, a naive distance-based approach \cite{KestemontSKKD16} and a compression-based approach \cite{Halvani2018}. BERT performs worse on the more difficult open splits. O2D2 performs surprisingly poor on the Open All test split, which may be due to its calibration step, since the performance on the development split  is much larger (80.6 vs 67.5). Evaluating BERT on the private PAN sets is slow due to access to CPU-only machines ($\approx$1200h on a machine powered by Intel Xeon E5 CPU with 8GB RAM memory). However, based on the scores of the O2D2 and baseline approaches, we expect it to perform similarly to the open test sets. 
\noindent\paragraph{Comparison on our splits.}
We fine-tune and evaluate the BERT-based models on the larger XL splits introduced in Sec.~\ref{sec: dataset} and compare them to the PAN 2021 winner, O2D2. 
We also report performances of two other models and their ensembles in the Supplementary Material~\ref{sec: apx_tab_siambert}: a Siamese model (siamBERT) and a domain-adapted BERT pretrained on the PAN 2020 corpus with the MLM objective, then fine-tuned on each split.
In Tab.~\ref{tab: pan2020_pretraining} we notice that BERT outperforms charBERT on all the splits over almost all the metrics. We expected charBERT to provide better contextual embeddings for rare words (like named entities), since they incorporate character n-grams into the embeddings. Though BERT may represent rare words noisily, it is sufficiently robust for the PAN 2020 corpus.

\subsection{Qualitative examples reveal biases}
We next focus on better understanding the models' predictions through explainable AI ~\cite{xai-survey} techniques. Inspecting the attention scores is a common method of explaining a model's prediction that has been called into question in recent years \cite{Pruthi2020,Serrano2019}. We therefore follow recent explainability results \cite{Bastings2020} and use the Integrated Gradients (IG) ~\cite{Sundararajan2017} method from the Captum library~\cite{Kokhlikyan2020} to reveal the individual importance of words.

We analyze BERT models fine-tuned on a closed and an open set, checking for potential biases arising from the dataset splitting process.
In Fig.~\ref{fig: xai_on_splits}, we show how important each word is in the authorship verification decision. For BERT trained on the Closed$_{XL}$ split (1st row per pair), the most important ones are the named entities. This initial focus is reduced when fine-tuning the model on the Open UF$_{XL}$ split (which keeps training and testing fandoms disjoint). This suggests that fandom-specific named entities encountered at test time are less likely to be exploited for the prediction, since they were not seen during training. Furthermore, the open validation splits help with generalization at the \emph{model selection} step. This is due to measuring the model's performance against fandom and author-specific information unseen at training time.

\subsection{Replacing named entities improves generalization}
We hypothesize that replacing the named entities may further help with generalization in a \emph{data-centric} fashion, prohibiting the model to exploit them at train time. To this end, we replace the named entities from the Open All XS dataset with their corresponding type (\eg \emph{Wolverine}$\to$\emph{person})\footnote{\url{https://spacy.io/api/entityrecognizer}}. We notice in Tab.~\ref{tab: pan20_xs} that this replacement step improves the \textit{overall} score for both models, strengthening our hypothesis about the role of named entities in authorship verification. Our results are in line with the previous works of \newcite{twitter_removal} and \newcite{attribution_removal}, which show that removing entities such as mentions, hashtags and topic information improves performance of authorship attribution.

Our results are further confirmed in a zero-shot scenario, under a significant distribution shift, when testing on the DarkReddit corpus introduced in Sec.~\ref{sec: darkreddit_dataset}. Specifically, we demonstrate in Tab.~\ref{tab: transfer} a significant performance gain when training without named entities. This suggests that the initial model was focusing on named entities in a spurious way.

\begin{table}[t]
\begin{tabular}{lcccc}
\toprule
                                    & \multicolumn{2}{c}{BERT}                & \multicolumn{2}{c}{charBERT}     \\ \cline{2-5} 
\multicolumn{1}{c}{\textbf{Metric}} & \textbf{w/ NE} & \textbf{w/o NE}        & \textbf{w/ NE} & \textbf{w/o NE} \\ \hline
\textit{F1}                         & \textbf{73.5}     & \multicolumn{1}{c|}{\textbf{73.5}} & 54.1              & \textbf{75.2}               \\
\textit{AUC}                        & 91.1              & \multicolumn{1}{c|}{\textbf{94.3}} & \textbf{89.0}     & 84.4                \\
\textit{F0.5}                       & 84.1              & \multicolumn{1}{c|}{\textbf{85.9}} & \textbf{72.9}     & 66.5               \\
\textit{C@1}                        & 78.1              & \multicolumn{1}{c|}{\textbf{78.7}} & 68.0              & \textbf{68.3}               \\
\textit{overall}                    & 81.7              & \multicolumn{1}{c|}{\textbf{83.1}} & 71.0              & \textbf{73.6}               \\ \hline
\end{tabular}
\caption{Performance of BERT and charBERT on the PAN Open All XS test split when using the raw dataset and hiding the named entities (w/o NE).}
\label{tab: pan20_xs}
\end{table}

\begingroup
\begin{table}[t]
    \setlength{\tabcolsep}{2pt} % Default value: 6pt
    \begin{center}
        \begin{tabular}{ l  ccccc }
            \toprule
            \shortstack{training set} &  \shortstack{\textit{F1}} &
            \shortstack{\textit{AUC}} &
            \shortstack{\textit{F0.5}} &
            \shortstack{\textit{C@1}} &
            \shortstack{\textit{overall}} \\
            
            \midrule
            Open All w/  NE & 69.5 & 83.0 & 58.9 & 56.4 & 67.0 \\
            Open All w/o NE & \textbf{74.1} & \textbf{86.4} & \textbf{64.4} & \textbf{65.4} & \textbf{72.6} \\
            \bottomrule
        \end{tabular}
    \end{center}
    \caption{Cross-corpus evaluation on DarkReddit. We compare the BERT models trained on the Open All XS dataset with and without named entities. Removing the named entities from the training set significantly improves the model's generalization across corpora.}
    \label{tab: transfer}
\end{table}
\endgroup

\section{Conclusions}
We introduced and published five splits of the PAN dataset ranging from the easiest \emph{closed} setup to increasingly more challenging settings. This enables a fine-grained evaluation and model selection. We showed that BERT-based baselines are competitive with top-scoring authorship verification methods and significantly outperform non-neural baselines.

Using Integrated Gradients, we showed that, distinctly from the closed split, the open splits help generalization at the \emph{model selection} step by preventing the model from overfitting on named entities of specific train authors or fandoms. We further improved generalization by replacing the named entities, making the models more robust to spurious features. This claim also holds under a strong distribution shift, when cross-evaluating the models on the significantly different DarkReddit dataset.

\section*{Acknowledgements}
This work has been supported in part by UEFISCDI, under Project PN-III-P2-2.1-PTE-2019-0532. Andrei Manolache was also supported by the International Max Planck Research School for Intelligent Systems (IMPRS-IS).

\section*{Limitations}

\noindent\paragraph{Closed vs. open splits.}
While our paper focuses on building more difficult open set splits, deploying authorship verification systems is application specific. This means that having methods trained on closed splits may be desirable in certain scenarios, such as when we are guaranteed that the test authors are known.

\noindent\paragraph{Noisy examples.} Collecting texts for building corpora for authorship verification can suffer from noisy data. Concretely, in both cases of PAN and DarkReddit, one user can write under multiple pseudonyms, leading to some different author examples to actually have the incorrect label. Moreover, multiple users can share the same account, leading to another issue where same author pairs are wrongly labeled. However, large-scale authorship verification models should be robust to this issue due to the large dataset size.

\noindent\paragraph{Long documents.} Our BERT-based baselines are capped at sequences of 512 tokens at most. This means that we can process at most 256 tokens from each text in a pair at a time. During training, we overcame this issues by selecting random chunks of texts. During evaluation, we aggregated predictions from all the chunks to obtain a prediction for the documents pair. This limits the representation power during both training and evaluating, due to encoding smaller contexts. Moreover, it slows down inference on longer examples, making it even more difficult to evaluate models on limited infrastructure. Further works should also include models that accommodate longer sequences.

\section*{Ethics Statement}
Authorship Verification systems may be deployed in non-ethical ways, by different organizations and parties, in order to track down vulnerable categories of people, such as journalists, dissidents, whistleblowers, etc. However, we believe that opening up research regarding authorship verification can help these vulnerable categories by raising awareness of the possibilities and limitations of state-of-the-art techniques and by mitigating their misuse. 

Our datasets are based on publicly available data and do not contain sensitive information.

\bibliography{anthology,custom}
\bibliographystyle{acl_natbib}

\appendix

\begin{table*}[t!]
    \setlength{\tabcolsep}{4pt} % Default value: 6pt
    \vspace{-0.1cm}
    \begin{center}
        \begin{tabular}{l  ccc | ccc | ccc | ccc}
            \toprule &
            \multicolumn{3}{c}{Closed$_{XL}$} & \multicolumn{3}{c}{Clopen$_{XL}$} & \multicolumn{3}{c}{Open UA$_{XL}$} & \multicolumn{3}{c}{Open UF$_{XL}$} \\
            \cmidrule(lr){2-4}
            \cmidrule(lr){5-7}
            \cmidrule(lr){8-10}
            \cmidrule(lr){11-13}
            \textbf{Metric} & 
            \textbf{sB} & \textbf{B}$^{\mathcal{\dagger}}$ & \textbf{B} &
            \textbf{sB} & \textbf{B$^{\mathcal{\dagger}}$} & \textbf{B} &
            \textbf{sB} & \textbf{B$^{\mathcal{\dagger}}$} & \textbf{B} &
            \textbf{sB} & \textbf{B$^{\mathcal{\dagger}}$} & \textbf{B}\\ 
            \midrule
            $F1$ & 85.0 & 94.4 & \textbf{95.0} & 79.8 & \textbf{96.8} & \textbf{96.8} & 84.1 & 84.0 & \textbf{89.2} & 83.1 & \textbf{91.3} & 90.8 \\
            $F0.5$& 84.8 & 93.5 & \textbf{94.5} & 80.2 & \textbf{97.3} & 97.0 & 85.3 & 86.9 & \textbf{88.1} & 84.1 & \textbf{91.2} & 88.8 \\
            $c@1$ & 86.0 & 93.9 & \textbf{94.5} & 81.4 & \textbf{96.6} & \textbf{96.6} & 85.5 & 83.7 & \textbf{88.2} & 84.2 & \textbf{90.8} & 90.0 \\
            $AUC$ & 93.1 & 98.4 & \textbf{98.6} & 89.6 & \textbf{99.5} & 99.4 & 92.7 & 92.4 & \textbf{95.3} & 92.0 & 96.8 & \textbf{96.9} \\
            $overall$ & 87.2 & 95.1 & \textbf{95.6} & 82.7 & \textbf{97.5} & 97.4 &  86.9 & 86.7 & \textbf{90.2} & 85.9 & \textbf{92.5} & 91.6 \\
            \bottomrule
        \end{tabular}
    \end{center}
    \caption{Comparison over large pre-trained models on PAN-2020 XL splits. BERT is very competitive and the domain-adapted BERT$^{\mathcal{\dagger}}$ does not consistently bring improvements over it. Distinctively from others, siamBERT never sees information from both documents at the same time, which significantly impacts its score.
    Note how the closed-set results (left side) are considerably higher than the open-set ones (right side), which might indicate the models overfit the styles of the known authors from the closed splits. We report all the PAN-2020 metrics. }
    \label{tab: other_models}
\end{table*}

\begin{table*}[t!]
    \setlength{\tabcolsep}{4.5pt} % Default value: 6pt
    \begin{center}
        \begin{tabular}{l  ccccc | ccccc}
            \toprule &
             \multicolumn{5}{c}{Open UA$_{XS}$} &
             \multicolumn{5}{c}{Open UF$_{XS}$} \\
            \cmidrule(lr){2-6}
            \cmidrule(lr){7-11}

            \textbf{Metric} & 
            \textbf{cB} & \textbf{sB} & \textbf{B$^{\mathcal{\dagger}}$} & \textbf{B} & \textbf{\shortstack{best\\ensemble}} & 
            \textbf{cB} & \textbf{sB} & \textbf{B$^{\mathcal{\dagger}}$} & \textbf{B} & \textbf{\shortstack{best\\ensemble}}\\
            \midrule
            $F1$ & 85.4 & 85.9 & 90.4 & 90.1 & 92.0 & 91.3 & 90.9 & 93.1 & 90.9 & 94.4 \\
            $AUC$ & 92.6 & 86.8 & 96.3 & 97.3 & 97.2 & 94.9 & 86.3 & 96.8 & 97.9 & 98.0 \\
            $overall$ & 87.0 & 87.2 & 92.1 & \textbf{93.5} & 93.4 & 91.2 & 88.9 & 93.2 & 93.2 & \textbf{95.3}\\
            \bottomrule
        \end{tabular}
    \end{center}
    \caption{Ensembling results on the XS Open UA and UF splits. We show that the individual models are complementary for the Open UF$_{XS}$ set, so the $overall$ score can be improved by combining them. However, this is not the case for Open UA$_{XS}$, indicating that the models might overfit on this split, most of them collapsing to a wrong prediction.}
    \label{tab: ensembles}
\end{table*}

\newpage 

\section{Other quantitative results}
\label{sec: apx_quant_results}

\subsection{SiamBERT and domain-adapted BERT}
\label{sec: apx_tab_siambert}
We list the performance of two other large pre-trained BERT-based models on the PAN XL  dataset splits in Tab.~\ref{tab: other_models}. The large gap between other models and siamBERT (\textbf{sB}) could be due to how the model functions, without learning over both documents simultaneously. BERT processes a pair of sequences, so the word-piece representations interact at every level before making a prediction based on the sequence pair embedding $h_{[CLS]}$. In contrast, siamBERT processes each sequence separately, making the word-pieces `interact' at the end through the sequence embeddings $h_{[CLS]}^{(1)}$ and $h_{[CLS]}^{(2)}$. The domain-adapted BERT (BERT$^{\mathcal{\dagger}}$) obtains similar results to BERT. Thus, the MLM fine-tuning step on Closed$_{XL}$ is not warranted, showing that adapting the representations to the domain of the downstream task brings no improvements. 

\subsection{Ensembling}
We measure the performance of various combinations over the previously described models. In Tab.~\ref{tab: ensembles}, we see how ensembling improves the performance on the Open UF$_{XS}$ set over the best model with over $2\%$. However, unexpectedly, the ensemble performance is weaker on the Open UA$_{XS}$ set. This
hurts the ensemble's robustness and might be a sign of overfitting, explained by having too many similar models that collapse to the same output, failing in the same points and overwriting the better prediction.

\section{Datasets}
\label{sec: apx_pan_splits}

\subsection{PAN dataset}
The PAN-2020 competition featured two datasets, a smaller one (52k pairs), intended for traditional shallow verification methods, and a larger one (275k pairs), intended for deep learning solutions. A document has an average of 21k words.\\

\noindent\textbf{PAN XL.} The large dataset has balanced classes (same vs different authors). Document pairs written by the same author always come from different fandoms (\eg Star Wars vs Harry Potter), while pairs written by different authors can belong to the same fandom or to different fandoms. Same author pairs are constructed from 41k authors, while different author pairs are constructed from 251k authors, with an overlap of 14k authors in both the same and different pairs. The XL dataset has 494k distinct documents that span 1.600 fandoms.\\

\noindent\textbf{PAN XS.} The small dataset is also balanced. Distinctly from the XL dataset, it has only cross-fandom pairs in both class pairs. This split allows fast prototyping through smaller experiments with models that have different components.

\subsection{Our splits}
We provide the construction details for all our splits below.

\paragraph{Closed split} In this setup, authors and fandoms at train time are also found in the validation and test sets (but with different documents). This split can hurt generalization, because it might work only on a subset of authors or even worse, on specific document pairs. Since we have no access to the PAN 2020 test set, we make the train, validation and test sets ourselves, by splitting the original pairs.
Each author pair $(a_i, a_j)$ in the DA pairs is unique, so splitting the DA pairs such that both test authors $a_i$ and $a_j$ are seen at train time is impossible. However, we relax this constraint and ensure that at least one of the authors in DA test pairs is seen at train time.

\paragraph{Clopen split} The Clopen split is similar to the closed split for the SA pairs. However, we remove the closed set constraint for the DA pairs and assign them randomly into train, validation and test. Thus, authors and fandoms in the Clopen test and validation sets might not be seen in the training set, making it a bit more general (more similar to the open sets). 

\paragraph{Open Unseen Authors split} In this split, authors from the test set should not appear in the training set. However, this is difficult to achieve strictly, so we split the PAN 2020 dataset into train and validation/test sets such that: i) authors of the SA test pairs do not appear in the SA train pairs; ii) some authors ($<5\%$) in the DA test pairs may appear in the DA train pairs; iii) most of the fandoms in the test set appear in the training set. 

\noindent\paragraph{Open Unseen Fandoms split} This split type has the following properties: i) fandoms in the validation/test sets are not seen during training; ii) some authors in the validation/test set may appear in the training set. To ensure no overlap between train and validation/test fandoms, training examples $(d_1, d_2, f_1, f_2)$ where either $f_1$ or $f_2$ appear in the validation/test fandoms are dropped. This results in approximately \textit{110K fewer train examples}.

\noindent\paragraph{Open all split} This split is the most difficult and the closest to the true open set setup in PAN 2021. Distinctly from the previous four splits, which were created using the original pairs, this split required sampling new document pairs and has the following properties: i) authors and fandoms in the test set have not been seen in the training data ii) authors in the validation set have not been seen in the training set, but the validation fandoms have been seen in the training set. 

\subsection{Distribution of named entities}
\label{sec: apx_pan_vs_reddit}
\begin{figure}[t!]
    \begin{center}
    \includegraphics[width=0.4\textwidth]{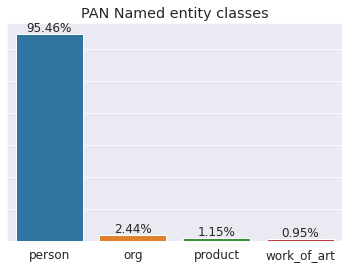}
    \end{center}
    \caption{
    Distribution of PAN named entities}
\label{fig: pan_dist}
\end{figure}

\begin{figure}[t]
    \begin{center}
    \includegraphics[width=0.4\textwidth]{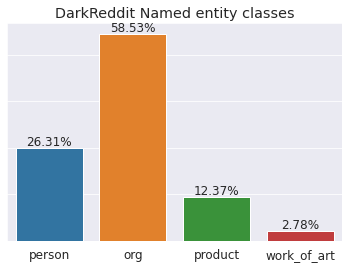}
    \end{center}
    \caption{
    Distribution of DarkReddit named entities.}
\label{fig: reddit_dist}
\end{figure}

We observe in Figures~\ref{fig: pan_dist} and~\ref{fig: reddit_dist} that the named entity distributions in the PAN and DarkReddit datasets are very different.

\begingroup
\setlength{\tabcolsep}{3.5pt} % Default value: 6pt
\begin{table*}[t!]

    \begin{center}
        \begin{tabular}{l ccccc | ccccc}
            \toprule
           &
          \multicolumn{5}{c}{\shortstack{Closed$_{XL}$ (split size)}}
          & \multicolumn{5}{c}{\shortstack{Clopen$_{XL}$ (split size)}}\\
          \cmidrule(lr){2-6}
          \cmidrule(lr){7-11}
          & \parbox[t]{8mm}{\multirow{2}{*}{\textbf{Total}}}
          & \multicolumn{2}{c}{\shortstack{\textbf{SA}}} 
          & \multicolumn{2}{c}{\shortstack{\textbf{DA}}}
          & \parbox[t]{8mm}{\multirow{2}{*}{\textbf{Total}}}
          & \multicolumn{2}{c}{\shortstack{\textbf{SA}}} 
          & \multicolumn{2}{c}{\shortstack{\textbf{DA}}}\\
            \cmidrule(lr){3-4}
            \cmidrule(lr){5-6}
            \cmidrule(lr){8-9}
            \cmidrule(lr){10-11}
            \multicolumn{1}{c}{\textbf{Split}} && SF & CF & SF & CF && SF & CF & SF & CF \\
            \toprule
            Train & 248,322 & 0 & 133,359 & 22,064 & 92,909 & 248,688 & 0 & 133,359 & 20,945 & 94,384\\
            Valid & 13,449 & 0 & 7,024 & 356 & 6,069 & 13,093 & 0 & 7,024 & 1,072 & 4,997 \\
            Test & 13,784 & 0 & 7,395 & 355 & 6,034& 13,784 & 0 & 7,395 & 1,114 & 5,275\\
            \bottomrule
        \end{tabular}
    \end{center}
    \label{tab: pan2020_xl_closed}
 \caption{PAN-2020 XL dataset - Closed-set splits, broken down into Same Author (SA) vs Different Author (DA). Each class is further divided into Same Fandom (SF) and Cross-Fandom (CF) pairs.}
\end{table*}
\endgroup

\begingroup
\setlength{\tabcolsep}{2pt} % Default value: 6pt
\begin{table*}[t!]
    \begin{center}
        \begin{tabular}{l cccc | cccc | cccc}
            \toprule
           &
          \multicolumn{4}{c}{\shortstack{Open UA$_{XL}$ (split size)}} 
          & \multicolumn{4}{c}{\shortstack{Open UF$_{XL}$ (split size)}}
          & \multicolumn{4}{c}{\shortstack{Open UAll$_{XL}$ (split size)}}\\
          \cmidrule(lr){2-5}
          \cmidrule(lr){6-9}
          \cmidrule(lr){10-13}
          & \multicolumn{2}{c}{\shortstack{\textbf{SA}}} 
          & \multicolumn{2}{c}{\shortstack{\textbf{DA}}}
          & \multicolumn{2}{c}{\shortstack{\textbf{SA}}} 
          & \multicolumn{2}{c}{\shortstack{\textbf{DA}}}
          & \multicolumn{2}{c}{\shortstack{\textbf{SA}}} 
          & \multicolumn{2}{c}{\shortstack{\textbf{DA}}}\\
            \cmidrule(lr){2-3}
            \cmidrule(lr){4-5}
            \cmidrule(lr){6-7}
            \cmidrule(lr){8-9}
            \cmidrule(lr){10-11}
            \cmidrule(lr){12-13}
            \multicolumn{1}{c}{\textbf{Split}} & SF & CF & SF & CF & SF & CF & SF & CF & SF & CF & SF & CF \\
            \toprule
            Train & 0 & 133,367 & 18,840 & 96,492 & 0 & 71,826 & 20,779 & 41,385 & 0 & 124,000 & 62,286 & 61,715\\
            Valid & 0 & 7,023 & 2,230 & 3,836 & 0 & 7,047 & 1,176 & 5,232 & 0 & 6,852 & 2,966& 3,885\\
            Test & 0 & 7,388 & 2,061 & 4,328 & 0 & 7,056 & 1,176 & 5,233 & 0 & 6,853 & 1,633& 5,218\\
            \bottomrule
        \end{tabular}
    \end{center}
    \label{tab: pan2020_xl_open}
    \caption{PAN-2020 XL - Open-set splits: Unseen Authors (UA$_{XL}$), Unseen Fandoms (UF$_{XL}$) and Unseen All (UAll$_{XL}$), broken down into Same Author (SA) vs Different Author (DA). Each class is further divided into Same Fandom (SF) and Cross-Fandom (CF) pairs.}
\end{table*}
\endgroup

\begin{figure*}[t!]
    \begin{center}
    \includegraphics[width=0.8\textwidth]{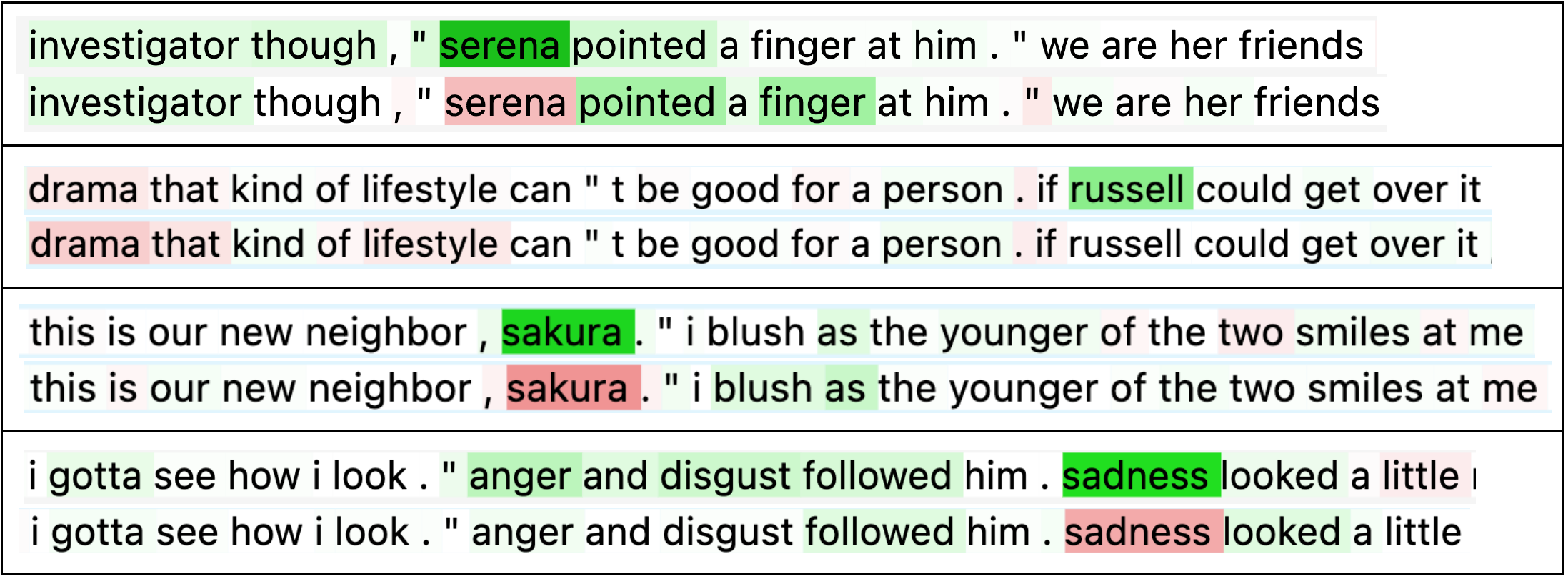}
    \end{center}
    \caption{
    Explainability analysis using Integrated Gradients for other samples, when fine-tuning BERT on Closed$_{XL}$ and Open UF$_{XL}$ respectively.}
\label{fig: xai_on_splits2}
\end{figure*}

\end{document}